\documentclass[sigconf,natbib=true,nonacm=true,anonymous=false]{acmart}
\usepackage{booktabs,multirow,xcolor}
\usepackage{graphicx}
\usepackage{subcaption}
\usepackage{makecell}
\usepackage{bm}
\AtBeginDocument{%
  }

\begin{document}

\title{CID-TKG: Collaborative Historical Invariance and Evolutionary \\ Dynamics Learning for Temporal Knowledge Graph Reasoning}


\author{Shuai-Long Lei}
\affiliation{%
  \institution{University of Science and Technology Beijing}
  \city{Haidian}
  \state{Beijing}
  \country{China}
}
\email{shuailong0lei@gmail.com}

\author{Xiaobin Zhu}
\authornote{Corresponding author}
\affiliation{%
  \institution{University of Science and Technology Beijing}
  \city{Haidian}
  \state{Beijing}
  \country{China}
}
\email{zhuxiaobin@ustb.edu.cn}

\author{Jiarui Liang}
\affiliation{%
  \institution{University of Science and Technology Beijing}
  \city{Haidian}
  \state{Beijing}
  \country{China}
}
\email{m202510664@xs.ustb.edu.cn}

\author{Guoxi Sun}
\affiliation{%
  \institution{University of Science and Technology Beijing}
  \city{Haidian}
  \state{Beijing}
  \country{China}
}
\email{sgxxyyds@foxmail.com}

\author{Zhiyu Fang}
\affiliation{%
  \institution{University of Science and Technology Beijing}
  \city{Haidian}
  \state{Beijing}
  \country{China}
}
\email{mr.fangzy@foxmail.com}

\author{Xu-Cheng Yin}
\affiliation{%
  \institution{University of Science and Technology Beijing}
  \city{Haidian}
  \state{Beijing}
  \country{China}
}
\email{xuchengyin@ustb.edu.cn}

\renewcommand{\shortauthors}{Trovato et al.}

\begin{abstract}
Temporal knowledge graph (TKG) reasoning aims to infer future facts at unseen timestamps from temporally evolving entities and relations. 
Despite recent progress, existing approaches still suffer from inherent limitations due to their inductive biases, as they predominantly rely on time-invariant or weakly time-dependent structures and overlook the evolutionary dynamics.
To overcome this limitation, we propose a novel collaborative learning framework for TKGR (dubbed CID-TKG) that integrates evolutionary dynamics and historical invariance semantics as an effective inductive bias for reasoning.
Specifically, CID-TKG constructs a historical invariance graph to capture long-term structural regularities and an evolutionary dynamics graph to model short-term temporal transitions. 
Dedicated encoders are then employed to learn representations from each structure.
To alleviate semantic discrepancies across the two structures, we decompose relations into view-specific representations and align view-specific query representations via a contrastive objective, which promotes cross-view consistency while suppressing view-specific noise. 
Extensive experiments verify that our CID-TKG achieves state-of-the-art performance under extrapolation settings.
\end{abstract}




\maketitle

\begin{figure}[t]
  \centering
  \includegraphics[width=\columnwidth]{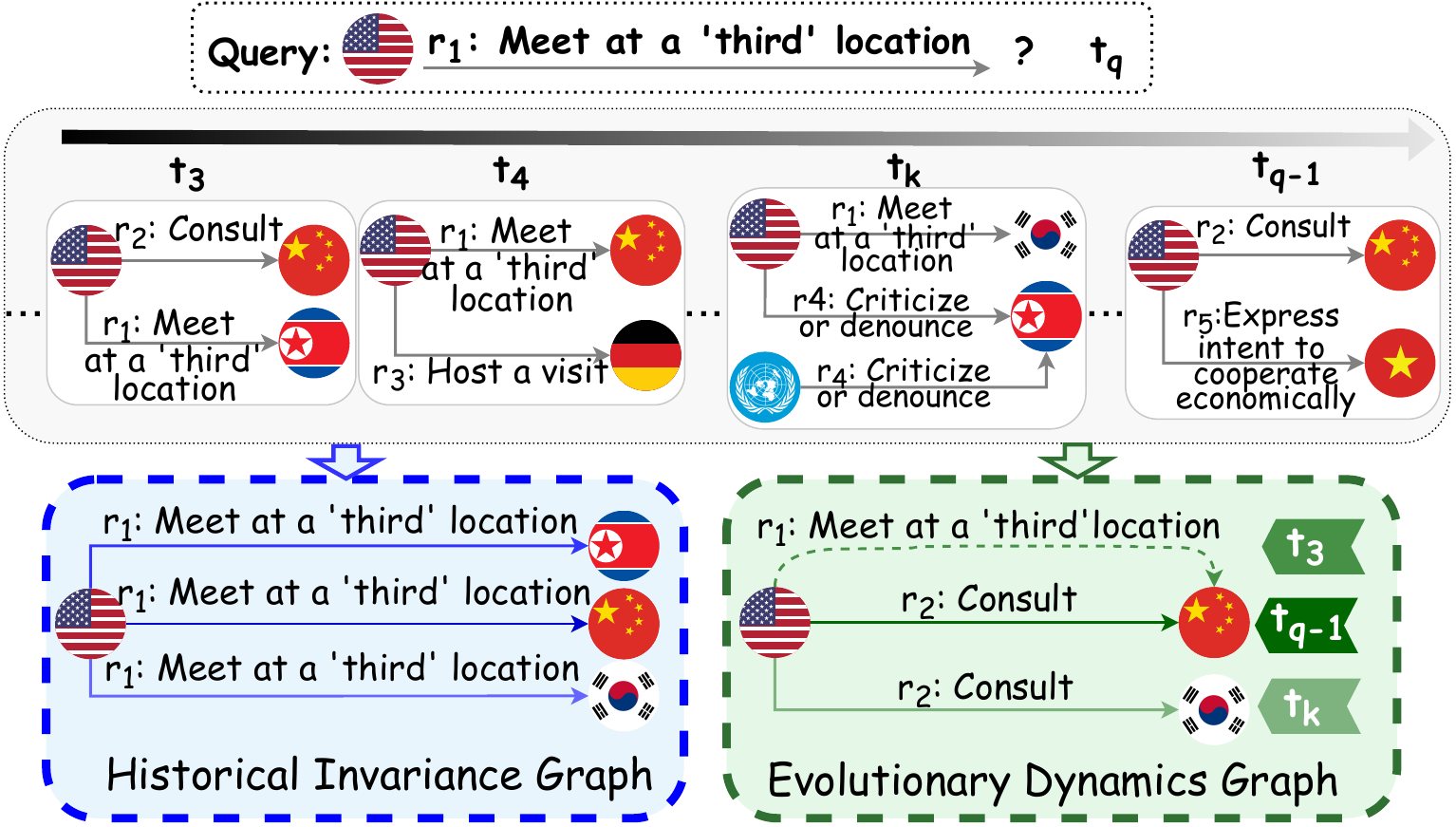}
  \caption{An illustrative example of CID-TKG on the ICEWS dataset.
  The blue region denotes the historical invariance graph, while the green region denotes the evolutionary dynamics graph.
  Darker colors indicate higher attention scores in both graphs.
  }
  \label{fig:problem}
  \Description{Illustration of CID-TKG on the ICEWS dataset, showing historical invariance and evolutionary dynamics graphs with attention intensities.}
\end{figure}

\section{INTRODUCTION}
Knowledge graphs (KGs) represent real-world knowledge as multi-relational directed graphs and support various downstream applications, such as recommendation \cite{tang2024editkg,wang2023mixed}, question answering \cite{atif2023beamqa,liu2022joint}, and information retrieval \cite{xu2024retrieval,ding2024enhancing}.
As many real-world facts are time-dependent, methods of temporal knowledge graphs (TKGs) explicitly model temporal information, where each fact is a quadruple $(subject, relation, object, timestamp)$ and organized as a sequence of time-ordered snapshots.
Due to the inherent incompleteness of TKGs \cite{zhang2025historically,du2025hawkes,chen2024local,liang2023learn}, temporal knowledge graph reasoning (TKGR) aims to infer missing facts at specific timestamps.
Existing TKGR methods can be broadly categorized into interpolation and extrapolation reasoning. 
Given an observed time range $[0,T]$, interpolation infers facts within this range, while extrapolation predicts future facts. 
As extrapolation reasoning is more challenging yet practically valuable, this work focuses on this paradigm.

A key challenge in predicting future events lies in effectively modeling the structures relevant to a given query. Recent methods typically adopt a hybrid paradigm, combining graph neural networks (GNNs) for spatial structure modeling with recurrent neural networks (RNNs) to capture temporal evolution, while incorporating time-invariant or weakly time-dependent auxiliary structures to facilitate reasoning.
Specifically, RE-GCN \cite{li2021temporal} constructs static attribute graphs to enrich entity semantics. CRNet \cite{wang2022crnet} builds candidate graphs to model interactions among future concurrent events. RETIA \cite{liu2023retia} introduces relation hypergraphs to alleviate relational information isolation. RPC \cite{liang2023learn} proposes relational correlation graphs to capture dependencies among relations. These methods primarily enhance entity- or relation-level representations through time-invariant auxiliary structures.
In addition, several approaches model weakly time-dependent structures by constructing global graphs over historical facts. For example, TiRGN \cite{li2022tirgn} builds a global historical graph to identify historical occurrences of queried quadruples, while HisRES \cite{zhang2025historically} constructs a global relevance graph to reduce the influence of distant events. LogCL \cite{fang2024arbitrary} further employs contrastive learning to fuse local semantics with global information. Despite their effectiveness, the above methods mainly rely on static or weakly time-dependent historical information and often ignore the evolutionary dynamics.

To address the above limitations, we transform a TKG into two complementary views: a historical invariance graph and an evolutionary dynamics graph. 
Fig.~\ref{fig:problem} illustrates this construction using an example from the ICEWS dataset. 
Given the query $q=$(\textit{North America}, \textit{Meet at a `third' location}, ?, $t_q$), we first collect all historical facts of the form (\textit{North America}, \textit{Meet at a `third' location}, $e_o$, t) with $t < t_q$. 
By removing timestamps and aggregating these facts, we construct the historical invariance graph (blue dashed region). 
In parallel, we aggregate temporally proximate and logically related recent events to form the evolutionary dynamics graph, capturing query-relevant temporal transitions. 
The relations \textit{Consult} and \textit{Meet\_at\_a\_`third'\_location} exhibit a temporal logical dependency, as identified by TLogic \cite{liu2022tlogic}.
As illustrated, the historical invariance graph provides stable and informative prior information due to recurring relational patterns, while the evolutionary dynamics graph captures recent, query-specific dynamics. These two complementary views jointly support more reliable tail entity prediction.

In this work, we propose CID-TKG, a novel collaborative learning framework that jointly models evolutionary dynamics and historical invariance structures through two complementary graph views. 
Specifically, we construct a historical invariance graph that captures globally relevant historical facts, and an evolutionary dynamics graph that models recent, logically related evolutionary events. 
The historical invariance graph provides stable prior information for prediction, while the evolutionary dynamics graph captures temporal dynamics, and the two graphs complement each other to enable accurate extrapolation reasoning.
Moreover, we introduce relation decomposition to explicitly separate relation representations across different graph views, alleviating semantic conflicts in multi-view learning. 
To further mitigate view-specific noise, CID-TKG employs a contrastive alignment constraint that aligns semantic representations across graph views, ensuring cross-view consistency and improving robustness. Our main contributions are four-fold:
\begin{itemize}
  \item We propose a novel collaborative learning framework for TKGR (dubbed CID-TKG) that integrates evolutionary dynamics and historical invariance semantics as an effective inductive bias for reasoning.
  \item To the best of our knowledge, CID-TKG is the first attempt in TKGR to explicitly separate and collaboratively learn invariant and dynamic structural patterns.
  \item We theoretically prove that the collaborative learning can reduce the lower bound of the minimum error, offering principled insights into the benefit of our CID-TKG.
  \item Extensive experiments on multiple benchmark datasets verify that our CID-TKG consistently achieves state-of-the-art performance in extrapolation settings.
\end{itemize}

\section{RELATED WORK}
\subsection{Interpolation Reasoning}
Interpolation reasoning focuses on inferring facts at observed past timestamps. Many existing methods aim to capture temporal dependencies through diverse modeling strategies \cite{leblay2018deriving,lacroixtensor,messner2022temporal,fang2024transformer,fang2024arbitrary,liu2025terdy,ying2024simple,li2023teast}. For example, TTransE \cite{leblay2018deriving} extends TransE \cite{bordes2013translating} by modeling time and relations as translation operations, while TComplex \cite{lacroixtensor} captures temporal dependencies while supporting atemporal facts. BoxTE \cite{messner2022temporal} incorporates temporal information via relation-specific transition matrices, and ECEformer \cite{fang2024transformer} models temporal patterns through evolutionary event chains.
However, these methods depend on explicit timestamp representations, which limits their generalization to unseen future timestamps and reduces their applicability to extrapolation reasoning.

\subsection{Extrapolation Reasoning}
\subsubsection{Embedding-Based TKGR}
Since TKGs are inherently graph-structured, many extrapolation reasoning methods adopt GNNs to learn entity and relation embeddings.
RE-GCN \cite{li2021temporal} pioneers this paradigm by modeling spatial structures with GNNs and capturing temporal evolution across snapshots using RNNs, inspiring a series of follow-up studies. 
CRNet \cite{wang2022crnet} constructs candidate graphs to model interactions among concurrent future events. 
Other approaches emphasize relation-centric modeling to facilitate entity representation learning \cite{chen2021dacha,zhang2023learning,liu2024temporal,liu2023retia,liang2023learn}, including DACHA \cite{chen2021dacha}, which captures correlations among historical relations, L$^2$TKG \cite{zhang2023learning}, which builds latent relation graphs, DHE-TKG \cite{liu2024temporal}, which introduces hypergraph structures to capture high-order interactions.
Overall, these methods primarily enrich entity and relation representations through time-invariant auxiliary structures.
Several methods further model weakly time-dependent structures by capturing the evolution of local events or constructing global graphs over historical facts.
TANGO \cite{han2021learning} models continuous-time dynamics using ordinary differential equations. HisMatch \cite{li2022hismatch}, CEN \cite{li2022complex}, and GHT \cite{sun2022graph} enhance sequential modeling using attention mechanisms, convolutional networks, or Transformers. 
TiRGN \cite{li2022tirgn} builds a global historical graph to model event recurrence, and subsequent methods such as LogCL \cite{chen2024local}, HisRES \cite{zhang2025historically}, and HGLS \cite{zhang2023learning_2} further capture long-term dependencies among historical events. 
DiMNet \cite{dong2025disentangled} disentangles entity representations to separately model long-term stability and short-term dynamics.
Few approaches introduce explicit time-aware mechanisms.
HERLN \cite{du2025hawkes} and HTCCN \cite{chen2024htccn} employ Hawkes processes to model temporal decay. TITer \cite{sun2021timetraveler} generates query-specific paths via reinforcement learning. 
However, these methods implicitly construct evolutionary dynamics structures and typically rely on limited relational correlations, restricting the ability to capture rich structural evolution patterns.

\subsubsection{Rule-Based TKGR}
In addition to embedding-based methods, another line of work addresses TKGR by learning temporal logical rules instead of entity embeddings. 
TLogic \cite{liu2022tlogic} extracts temporal logical rules via non-increasing temporal random walks, estimates rule confidence, and derives predictions through noisy-OR aggregation. 
Subsequent methods extend this framework. 
TR-Rules \cite{li2023tr} mines temporal acyclic rules and introduces window-based confidence to mitigate temporal redundancy, while TempValid \cite{huang2024confidence} further incorporates temporal decay into rule confidence estimation.
Unlike TLogic and its variants, INFER \cite{li2025infer} incorporates temporal validity, fact frequency, and embedding information for extrapolation reasoning, while DaeMon \cite{dong2023adaptive} learns continuous and implicit path representations through neural networks without explicitly constructing logical rules.
Overall, rule-based methods primarily capture temporal dependencies through rule patterns, while largely overlooking stable semantic information that persists across time.

\subsubsection{LLM-Based TKGR}
Recently, several studies explore leveraging the inductive and reasoning capabilities of large language models (LLMs) for TKGR. 
ICL \cite{lee2023temporal} applies in-context learning by prompting LLMs with historical inputs, while GenTKG \cite{liao2024gentkg} and LLM-DA \cite{wang2024large} incorporate temporal logical rules to retrieve and refine historical sequences.
CoH \cite{xia2024chain} leverages LLMs to explore high-order historical chains and combines their semantic reasoning with structural predictions from graph models.
G2S \cite{bai2025g2s} adopts a two-stage framework that combines ID-based quadruples with textual representations, and AnRe \cite{tang2025anre} further enhances temporal reasoning by guiding LLMs with few-shot analogical examples.
Overall, LLM-based approaches mainly rely on the reasoning and generalization capabilities of LLMs, with limited emphasis on explicitly modeling temporal structures and persistent semantic information.

\begin{table}[h]
  \centering
  \caption{Summary of Important Notations}
  \small
  \begin{tabular}{l p{6.5cm}}
  \toprule
  \textbf{Notations} & \textbf{Descriptions} \\
  \midrule
  $\mathcal{E}, \mathcal{R}, \mathcal{T}, \mathcal{F}$ 
  & Entity, relation, timestamp, and temporal fact set in a TKG. \\

  $\mathcal{G}_t$ 
  & Temporal knowledge graph snapshot at timestamp $t$. \\

  $\mathbf{e}_{s,t}, \mathbf{r}_{t}, \mathbf{e}_{o,t}$ 
  & Subject, relation, and object embeddings at timestamp $t$. \\

  $\mathbf{E}_t, \mathbf{R}_t$ 
  & Entity and relation embedding matrices at timestamp $t$. \\

  $\mathcal{G}^{D}, \mathcal{G}^{I}$ 
  & Evolutionary dynamics and historical invariance graph. \\

  $\mathbf{e}^{D}, \mathbf{e}^{I}$ 
  & Entity embeddings of both graphs. \\

  $\mathbf{r}^{D}, \mathbf{r}^{I}$ 
  & Relation embeddings of both graphs. \\

  $\mathbf{E}^{D}, \mathbf{E}^{I}$ 
  & Entity embedding matrices of both graphs. \\

  $\mathbf{R}^{D}, \mathbf{R}^{I}$ 
  & Relation embedding matrices of both graphs. \\

  $d$ 
  & Embedding dimensionality. \\

  $L$
  & Number of historical snapshots used for modeling. \\

  $\mathcal{Q}_{t_q}$
  & Set of prediction queries associated with timestamp $t_q$. \\
  \bottomrule
  \end{tabular}
\end{table}

\section{Preliminary}
\subsection{\bf Task Definition.}
A temporal knowledge graph (TKG) is defined as $(\mathcal{E}, \mathcal{R}, \mathcal{T}, \mathcal{F})$, where $\mathcal{E}$, $\mathcal{R}$, $\mathcal{T}$, and $\mathcal{F}$ denote the sets of entities, relations, timestamps, and temporal facts, respectively. 
Each fact $f \in \mathcal{F}$ is a quadruple $(e_s, r, e_o, t)$ with $e_s, e_o \in \mathcal{E}$, $r \in \mathcal{R}$, and $t \in \mathcal{T}$. 
Facts are organized into time-ordered snapshots $\mathcal{G}=\{\mathcal{G}_1, \mathcal{G}_2, \dots\}$, where $\mathcal{G}_t=\{(e_s, r, e_o, t)\mid (e_s, r, e_o, t)\in\mathcal{F}\}$.
Extrapolation reasoning aims to infer facts at unseen future timestamps and is formulated as a link prediction task that answers queries $q=(e_s, r, ?, t_q)$. During training, only historical facts with $t_i < t_q$ are observable, while facts at timestamp $t_q$ and beyond remain unobserved.
We denote by $\mathcal{Q}_{t_q}$ the set of such queries at timestamp $t_q$.
Following prior work, inverse relations are introduced by adding edges $(e_o, r^{-1}, e_s, t)$, allowing subject prediction queries to be transformed into tail prediction queries. 
For simplicity, we describe the task in terms of tail entity prediction in the subsequent analysis.

\subsection{\bf Temporal Logical Rule.}
A temporal logical rule $\mathcal{TR}$ is defined as
\begin{equation}
(A, r_h, B, t_2) \leftarrow (A, r_b, B, t_1), \quad t_2 > t_1,
\end{equation}
indicating that the occurrence of relation $r_b$ between entities $A$ and $B$ at time $t_1$ supports the occurrence of relation $r_h$ between the same entities at a later time $t_2$. The confidence of a rule is measured by how often the rule
body is followed by the rule head.
In this work, we leverage temporal logical rules to construct the evolutionary dynamics graph, rather than focusing on rule mining itself. We therefore adopt TLogic \cite{liu2022tlogic} to mine cyclic temporal logical rules from the observed TKG.

\begin{figure*}[t!]
  \centering
  \includegraphics[width=\textwidth]{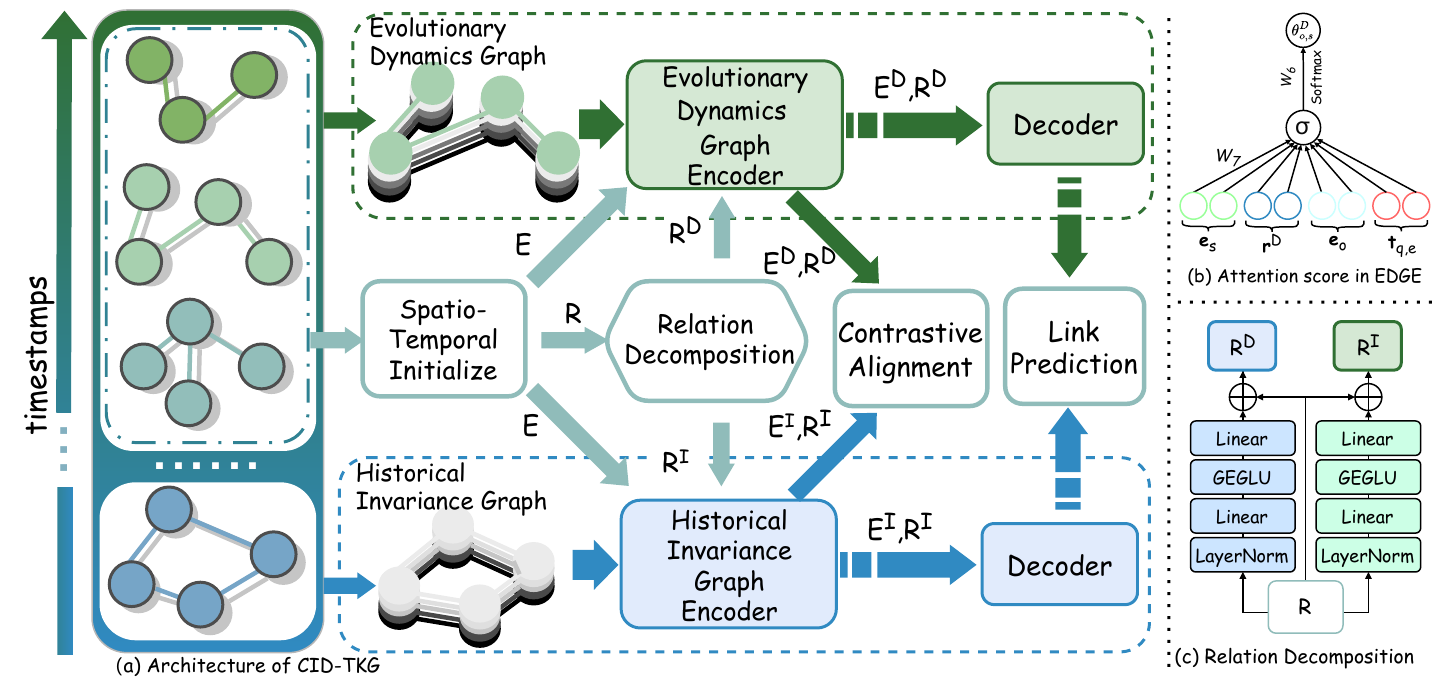}
  \vspace{-16pt}
  \caption{Overall architecture of CID-TKG (a). The framework consists of
  spatio-temporal initialization, relation decomposition (c),
  a historical invariance graph encoder, and an evolutionary dynamics graph encoder
  with time-aware attention scores (b), followed by contrastive
  alignment for collaborative representation learning.
  }
  \label{fig:pipeline}
\end{figure*}

\section{METHODOLOGY}
\subsection{Theoretical Analysis}
Let $Y\in\mathcal E$ denote the target tail entity and let $Z\in\mathcal Z$ be the representation used for prediction.
A predictor is a measurable function $\phi:\mathcal Z\rightarrow\mathcal E$.
Under the $0$--$1$ loss, the prediction error of $\phi$ is defined as:
\begin{equation}
P_e(\phi)=\Pr\big(\phi(Z)\neq Y\big).
\end{equation}

The Bayes optimal error given representation $Z$ is defined as:
\begin{equation}
R^*(Z)
=
\inf_{\phi:\mathcal Z\rightarrow\mathcal E}
\Pr\big(\phi(Z)\neq Y\big).
\end{equation}
Under the $0$--$1$ loss, the infimum is attained by the Bayes (MAP) predictor; hence, $R^*(Z)$ corresponds to the minimum achievable error among all predictors that rely solely on the representation $Z$.
Let $|\mathcal E|=M$.
According to Fano's inequality \cite{fano1961transmission}, the Bayes optimal error is lower bounded as:
\begin{equation}
R^*(Z)
\ge
\frac{H(Y)-I(Y;Z)-1}{\log M},
\end{equation}
where $H(\cdot)$ denotes entropy and $I(\cdot;\cdot)$ denotes mutual information.
This inequality suggests that tightening the achievable-error lower bound can be achieved by increasing the mutual information $I(Y;Z)$ between the representation and the target entity.
Most existing approaches rely on time-independent or weakly time-dependent structures in TKGs, which we denote as $\mathcal{G}^{I}$. Let $Z^I=f_{\theta^I}(\mathcal{G}^{I})$ be the resulting representation, yielding mutual information $I(Y;Z^I)$. However, $Z^I$ may fail to capture complementary information encoded in fully evolutionary dynamics structures.
To incorporate complementary information, we consider the evolutionary dynamics graph $\mathcal{G}^{D}$ with representation $Z^D=f_{\theta^D}(\mathcal{G}^{D})$, and form a collaborative representation:
\begin{equation}
Z_{\mathrm{collab}}=(Z^{I}, Z^{D}).
\end{equation}
By the chain rule of mutual information, we have
\begin{equation}
I(Y;Z_{\mathrm{collab}})
=
I(Y;Z^{I})
+
I(Y;Z^{D} \mid Z^{I}).
\end{equation}
When the dynamic view provides non-redundant information beyond $\mathcal{G}^{I}$, i.e.,
\begin{equation}
I(Y;Z^{D} \mid Z^{I})>0,
\end{equation}
it follows that $I(Y;Z_{\mathrm{collab}})>I(Y;Z^{I})$, thereby yielding a strictly tighter Fano lower bound on the Bayes optimal error.
This analysis provides an information-theoretic motivation for augmenting time-independent or weakly time-dependent representations with evolutionary dynamics, highlighting that dynamic representations are effective only when they contribute non-redundant information beyond the invariant structure, i.e., when $I(Y;Z^{D}\mid Z^{I})>0$.

Motivated by this insight, CID-TKG adopts a collaborative learning framework that jointly models historical invariance and evolutionary dynamics. We introduce relation decomposition to disentangle heterogeneous relational semantics across views, and employ contrastive learning to align the two representations, suppressing spurious temporal variations while preserving relation-consistent patterns. These designs promote a larger $I(Y;Z^{D}\mid Z^{I})$, leading to a tighter Fano lower bound and improved extrapolation performance.


\subsection{Model Overview}
From the above theoretical analysis, we propose CID-TKG (illustrated in Fig.~\ref{fig:pipeline}) for entity extrapolation in TKGs. CID-TKG collaboratively learns historical invariance and evolutionary dynamics from observed facts by constructing a historical invariance graph and an evolutionary dynamics graph, respectively. Recent subgraph facts are incorporated to initialize the spatio-temporal structure and facilitate collaborative learning.
To capture heterogeneous relational semantics, relation representations are decomposed into view-specific components. 
Entity and decomposed relation embeddings are then encoded by the Historical Invariance Graph Encoder (HIGE) and the Evolutionary Dynamics Graph Encoder (EDGE) to learn stable structural patterns and temporal dynamics. 
To suppress view-specific noise and enforce cross-view consistency, we adopt a contrastive alignment mechanism to align representations of the two structures. 
Finally, invariant and dynamic representations are combined in the decoder for entity prediction. 

\subsection{Collaborative Graph Construction}
As discussed in the Introduction and Preliminaries, most existing extrapolation methods primarily rely on historical invariance structures. 
While such structures provide stable semantic priors, they often lack sufficient evolutionary evidence to characterize recent dynamics for future prediction. 
Conversely, modeling recent dynamics alone is also insufficient, as it benefits from historical invariance structures that offer reliable semantic references and long-term regularities.
Motivated by these observations, CID-TKG collaboratively constructs a historical invariance graph and an evolutionary dynamics graph. 
Compared with the historical invariance graph, the evolutionary dynamics graph captures additional query-specific temporal signals, providing complementary information beyond time-invariant historical structures. 
Together, these two graphs enable the model to capture complementary temporal knowledge from stable structures and evolving patterns.

\subsubsection{Historical Invariance Graph}
We construct the historical invariance graph by aggregating historical events while discarding explicit temporal information. To reduce noise, only historical facts sharing the same relation as the query are retained.
Specifically, given a query $q=(e_s, r, ?, t_q)$, we retrieve all historical facts
$f^{I}_{t_q}=(e_s, r, e_o, t)$ with $t<t_q$ and incorporate them into the corresponding historical invariance subgraph $G^{I}_{t_q}\in\mathcal{G}^{I}$. 
As $t_q$ increases, the historical invariance graph and the evolutionary dynamics graph are constructed from an expanded set of historical facts observed prior to $t_q$.
For each timestamp $t_q$, all facts associated with its queries are aggregated into the same invariant and dynamic subgraphs, denoted as $G^{I}_{t_q}$ and $G^{D}_{t_q}$.

\subsubsection{Evolutionary Dynamics Graph}
The evolutionary dynamics graph aims to capture historical events that are temporally and logically close to the query $q=(e_s, r, ?, t_q)$. Thus, we adopt a temporal logical rule--based retrieval strategy to construct the evolutionary dynamics subgraph $G^{D}_{t_q}$.
Specifically, we first mine temporal logical rules using TLogic \cite{liu2022tlogic} and sort them by confidence in descending order. 
Given a query, the query relation $r$ is treated as the head relation $r_h$, and rules are selected accordingly. 
For each selected rule with body relation $r_b$, we retrieve historical facts of the form $f^{D}_{t_q}=(e_s, r_b, e_b, t)$ with $t<t_q$ in reverse chronological order, and incorporate them into $G^{D}_{t_q}$ until a predefined upper bound $N$ is reached.
In practice, we restrict rule retrieval to one-hop rules, which generally provide higher confidence and sufficient coverage, and skip rule--based retrieval for relations not covered by any mined rule.
This strategy enables the evolutionary dynamics graph to aggregate events that are both temporally proximate and logically related to the query. 
For each edge, we explicitly encode the relative time interval $\Delta t=t_q-t$ as edge information, allowing the model to capture fine-grained temporal evolution patterns.

\subsection{Initialization}

The initialization stage provides informative spatio-temporal representations to facilitate collaborative learning by leveraging temporal and structural information from historical subgraphs. Following \cite{li2022hismatch,zhang2025historically,chen2024local}, we incorporate time encoding to model temporal evolution and periodic event patterns.
Given the relative timestamp $\tau=t_q-t_i$, we compute a cosine-based temporal encoding:
\begin{equation}
\phi(\tau)=\cos\!\left(w_\tau \tau + b_\tau\right),
\end{equation}
which is concatenated with the entity embedding and projected as:
\begin{equation}
\mathbf{e}_t = W_0 \bigl[ \mathbf{e}_t \,\Vert\, \phi(\tau) \bigr],
\end{equation}
where $w_\tau$ and $b_\tau$ are learnable parameters, $[\cdot\Vert\cdot]$ denotes concatenation, and $W_0\in\mathbb{R}^{d\times 2d}$ is a linear projection matrix.

At each timestamp, we employ a graph convolutional network (GCN) to aggregate multi-relational neighborhood information within each snapshot:
\begin{equation}
\mathbf{e}_{o,t-1}^{\,l+1}
=
\sigma\!\left(
\sum_{(e_s,r,e_o)\in\mathcal{G}_{t-1}}
\frac{1}{c_o}
\, W_1^{\,l}
\kappa(\mathbf{e}_{s,t-1}^{\,l}, \mathbf{r}_{t-1}^{\,l})
+
W_2^{\,l} \mathbf{e}_{o,t-1}^{\,l}
\right),
\end{equation}
where $\mathbf{e}_{s,t-1}^{\,l}$ and $\mathbf{e}_{o,t-1}^{\,l}$ are entity embeddings at time $t\!-\!1$ in the $l$-th layer, $W_1^{\,l}$ and $W_2^{\,l}$ are learnable parameters, $c_o$ denotes the in-degree normalization factor, and $\sigma(\cdot)$ is the RReLU activation function. 
Following \cite{zhang2025historically}, relation embeddings are updated jointly in the GCN, and $\kappa(\cdot)$ denotes a one-dimensional convolution operator as in \cite{li2022tirgn}.

After modeling snapshot-level spatial structures with GCNs, we employ Gated Recurrent Units (GRUs) to capture temporal dependencies across snapshots. An entity-oriented GRU updates as:
\begin{equation}
\mathbf{E}_t = \mathrm{GRU}(\mathbf{E}_{t-1}, \mathbf{E}_{t-1}^{\mathrm{GCN}}),
\end{equation}
where $\mathbf{E}_t$ and $\mathbf{E}_{t-1}$ denote the entity embedding matrices at time steps $t$ and $t-1$, and $\mathbf{E}_{t-1}^{\mathrm{GCN}}$ is the GCN output at time $t-1$.
Relation representations are updated similarly using a relation-oriented GRU. An intermediate relation representation is first constructed as:
\begin{equation}
\mathbf{r}_t' = [\, \mathrm{pool}(\mathbf{E}_{r,t}) \,\Vert\, \mathbf{r} \,],
\end{equation}
where $\mathbf{E}_{r,t}$ denotes entities associated with relation $r$ at time $t$, $\mathrm{pool}(\cdot)$ denotes mean pooling. The intermediate representations are aggregated into $\mathbf{R}_t'$, the relation embeddings are updated by:
\begin{equation}
\mathbf{R}_t = \mathrm{GRU}(\mathbf{R}_{t-1}, \mathbf{R}_t').
\end{equation}

\subsection{Relation Decomposition}
In the collaborative graph setting, relations play different semantic roles in the historical invariance graph and the evolutionary dynamics graph. Specifically, relations in the evolutionary dynamics graph encode temporal transition patterns, while those in the historical invariance graph capture stable, time-agnostic associations. Using a shared relation embedding across both graphs may therefore lead to semantic interference.
To address this issue, we introduce a relation decomposition mechanism that learns view-specific relation representations. Formally, given a base relation embedding $r$, we derive two view-specific embeddings via multiplicative modulation:
\begin{align}
  \hat{r}^{I} &= \bigl( \mathbf{1} + g^{I}(r) \bigr) \odot r, \label{eq:re_s}\\
  \hat{r}^{D} &= \bigl( \mathbf{1} + g^{D}(r) \bigr) \odot r, \label{eq:re_d}
\end{align}
where $\hat{r}^{I}$ and $\hat{r}^{D}$ are used in the historical invariance graph and the evolutionary dynamics graph, respectively, $\mathbf{1}$ denotes an all-ones vector, and $\odot$ denotes element-wise multiplication.
Each view-specific function $g(\cdot)$ is implemented as:
\begin{equation}
  g(\cdot)
  =
  W_3
  \,
  \mathrm{Drop}\!\left(
  \mathrm{GEGLU}(\mathrm{LN}(\cdot))
  \right),
\label{eq:mlp}
\end{equation}
where $W_3$ is a learnable projection matrix, $\mathrm{LN}(\cdot)$ denotes layer normalization, $\mathrm{GEGLU}(\cdot)$ \cite{shazeer2020glu} applies a gated linear unit with GELU activation, and $\mathrm{Drop}(\cdot)$ denotes dropout. This decomposition mitigates representational interference between graph structures and encourages the evolutionary dynamics to contribute non-redundant information conditioned on historical invariance.

\begin{table*}[t!]
  \centering
  \caption{Statistics of datasets}
  \begin{tabular}{lcccccc}
    \toprule
  Dataset & Entities $|E|$ & Relations $|R|$ & Training Facts & Validation Facts & Testing Facts & Time granularity \\
  \hline
  ICEWS14s    & 6,869  & 230 & 74,845  & 8,514  & 7,371  & 24 hours \\
  ICEWS18    & 23,033 & 256 & 373,018 & 45,995 & 49,545 & 24 hours \\
  ICEWS05-15 & 10,094 & 251 & 368,868 & 46,302 & 46,159 & 24 hours \\
  GDELT      & 7,691  & 240 & 1,734,399 & 238,765 & 305,241 & 15 mins \\
  \bottomrule
  \end{tabular}
  \label{tab:dataset_statistics_horizontal}
  \end{table*}

\subsection{Graph Encoder}
To capture event correlations in both graphs, we adopt a Graph Attention Network (GAT) architecture with two view-specific encoders: the Historical Invariance Graph Encoder (HIGE) and the Evolutionary Dynamics Graph Encoder (EDGE). 
Entity and relation representations are initialized with spatio-temporal and decomposed embeddings, respectively.

\paragraph{\textbf{Historical Invariance Graph Encoder.}}
The attention coefficient in the HIGE is computed as:
\begin{equation}
  \theta^{I,l}_{o,s}
  =
  \frac{
  \exp\!\big(
  W^{l}_{4}\,
  \sigma\!\big(
  W^{l}_{5}\,
  [\,\mathbf{e}^{l}_{s} \Vert \hat{\mathbf{r}}^{I,l} \Vert \mathbf{e}^{l}_{o}\,]
  \big)
  \big)
  }{
  \sum_{(e_s',r') \in \mathcal{N}^{I}(o)}
  \exp\!\big(
  W^{l}_{4}\,
  \sigma\!\big(
  W^{l}_{5}\,
  [\,\mathbf{e}^{\prime l}_{s} \Vert \hat{\mathbf{r}}^{\prime I,l} \Vert \mathbf{e}^{l}_{o}\,]
  \big)
  \big)
  } .
\end{equation}

\paragraph{\textbf{Evolutionary Dynamics Graph Encoder.}}
To model temporal relevance, we encode the relative time interval between the query
$q=(e_s,r,?,t_q)$ and a historical fact $(e_s,r,e_o,t_e)$ as:
\begin{equation}
t_{q,e}=\Phi(t_q-t_e),
\end{equation}
where $\Phi(\cdot)$ follows TGAT \cite{xu2020inductive}:
\begin{equation}
\Phi(t)=\sqrt{\tfrac{1}{d}}\,[\cos(w_1 t+p_1),\ldots,\cos(w_d t+p_d)] .
\end{equation}
The attention coefficient in the EDGE is then defined as:
\begin{equation}
  \theta^{D,l}_{o,s}
  =
  \frac{
  \exp\!\big(
  W^{l}_{6}\,
  \sigma\!\big(
  W^{l}_{7}\,
  [\,\mathbf{e}^{l}_{s} \Vert \hat{\mathbf{r}}^{D,l} \Vert \mathbf{e}^{l}_{o} \Vert t_{q,e}\,]
  \big)
  \big)
  }{
  \sum_{(e_s',r') \in \mathcal{N}^{D}(o)}
  \exp\!\big(
  W^{l}_{6}\,
  \sigma\!\big(
  W^{l}_{7}\,
  [\,\mathbf{e}^{\prime l}_{s} \Vert \hat{\mathbf{r}}^{\prime D,l} \Vert \mathbf{e}^{l}_{o} \Vert t_{q,e}\,]
  \big)
  \big)
  } .
\end{equation}

\paragraph{\textbf{Message Aggregation.}}
For both graph views $v\in\{I,D\}$, entity representations are updated by:
\begin{equation}
  \mathbf{e}^{\,v,l+1}_{o}
  =
  \sigma\!\left(
  \sum_{(e_s,r,e_o)\in \mathcal{G}^{v}_{t_q}}
  \theta^{\,v,l}_{o,s}\,
  W^{\,v,l}_{8}\,
  \psi\!\left(\mathbf{e}^{\,v,l}_{s} + \hat{\mathbf{r}}^{\,v,l}\right)
  +
  W^{\,v,l}_{9}\,
  \mathbf{e}^{\,v,l}_{o}
  \right).
\end{equation}
Here, $\mathbf{e}^{l}_{s}$ and $\mathbf{e}^{l}_{o}$ denote the embeddings of the source and target entities at layer $l$, $\hat{\mathbf{r}}^{v,l}$ denotes the decomposed relation embedding for view $v$, and $\mathcal{N}^{v}(o)$ denotes the incoming neighbors of $e_o$ in graph view $v$.
$W^{(\cdot)}$ are learnable projection matrices, $\psi(\cdot)$ denotes a nonlinear transformation, and $\sigma(\cdot)$ is an activation function.

\subsection{Contrastive Alignment}
Through EDGE and HIGE, we obtain entity and relation representations from the evolutionary dynamics graph and the historical invariance graph, respectively.
When learned independently, these representations may contain view-specific noise.
To promote cross-view consistency, we introduce contrastive alignment as a collaborative constraint.

Given a query $q=(e_s,r,?,t_q)\in\mathcal{Q}_{t_q}$, we construct query representations under both views. Let $\mathbf{e}^{D}_{s}$ and $\mathbf{e}^{I}_{s}$ denote the representations of $e_s$, and $\mathbf{r}^{D}$ and $\mathbf{r}^{I}$ the corresponding decomposed relation embeddings. 
The query representations are defined as:
\begin{align}
\mathbf{z}^{D}_q &= \mathrm{MLP}\big([\mathbf{e}^{D}_{s} \Vert \mathbf{r}^{D}]\big),\\
\mathbf{z}^{I}_q &= \mathrm{MLP}\big([\mathbf{e}^{I}_{s} \Vert \mathbf{r}^{I}]\big),
\end{align}
where the MLP follows Eq.~(\ref{eq:mlp}) with independent parameters.
We align the two views using an InfoNCE-based contrastive loss as:
\begin{equation}
\mathcal{L}_{D \rightarrow I}
=
- \mathbb{E}_{q \in \mathcal{Q}_{t_q}}
\log
\frac{
\exp\!\left( \mathbf{z}^{D}_q \cdot \mathbf{z}^{I}_q / \gamma \right)
}{
\sum_{q' \in \mathcal{Q}_{t_q}}
\exp\!\left( \mathbf{z}^{D}_q \cdot \mathbf{z}^{I}_{q'} / \gamma \right)
},
\end{equation}
where $\gamma$ is a temperature parameter. The final contrastive objective is defined symmetrically as
$\mathcal{L}_{\mathrm{CoA}}=\mathcal{L}_{D \rightarrow I}+\mathcal{L}_{I \rightarrow D}$.
This contrastive alignment suppresses view-specific noise and encourages the evolutionary dynamics representation to contribute non-redundant information conditioned on the historical invariance structure.

\begin{table*}[]
  \centering
  \small
  \setlength{\tabcolsep}{3pt}
  \renewcommand{\arraystretch}{1.1}
  \caption{Performance of TKG entity extrapolation on ICEWS14s, ICEWS18, ICEWS05-15 and GDELT. The time-filtered MRR, Hits@1, Hits@3, and Hits@10 metrics are multiplied by 100. The best results are marked in \textbf{bold} and the second results are \underline{underlined}. We use $\clubsuit$, $\diamondsuit$, and $\spadesuit$ to denote path-based, LLM-based, and embedding-based methods, respectively.}
  \label{tab:main-results}
  \resizebox{0.97\textwidth}{!}{
  \begin{tabular}{c l *{4}{c} *{4}{c} *{4}{c} *{4}{c}}
  \toprule
  \multirow{2}{*}{Type} & \multirow{2}{*}{Model}
  & \multicolumn{4}{c}{ICEWS14s}
  & \multicolumn{4}{c}{ICEWS18}
  & \multicolumn{4}{c}{ICEWS05-15}
  & \multicolumn{4}{c}{GDELT} \\
  \cmidrule(lr){3-6}\cmidrule(lr){7-10}\cmidrule(lr){11-14}\cmidrule(lr){15-18}
  & & MRR & Hits@1 & Hits@3 & Hits@10
      & MRR & Hits@1 & Hits@3 & Hits@10
      & MRR & Hits@1 & Hits@3 & Hits@10
      & MRR & Hits@1 & Hits@3 & Hits@10 \\
  \midrule
  
  \multirow{3}{*}{$\clubsuit$} 
  & TLogic   & 43.04 & 33.56 & 48.27 & 61.23 & 29.82 & 20.54 & 33.95 & 48.53 & 46.97 & 36.21 & 53.13 & 67.43 & -- & -- & -- & -- \\
  & TR-Rules & 43.32 & 33.96 & 48.55 & 61.17 & 30.41 & 21.10 & 34.58 & 48.92 & 47.64 & 37.06 & 53.80 & 67.57 & -- & -- & -- & -- \\
  & INFER    & 44.46 & 35.03 & 49.37 & 62.31 & 32.22 & 22.39 & 36.41 & 51.52 & 48.73 & 38.32 & 54.61 & 68.48 & -- & -- & -- & -- \\
  \midrule
  
  \multirow{3}{*}{$\diamondsuit$} 
  & GenTKG & --    & 36.85 & 47.95 & 53.50 & --    & 24.25 & 37.25 & 42.10 & --    & --    & --    & --    & --    & --    & --    & --    \\
  & LLM-DA & 47.10 & 36.90 & 52.60 & 67.10 & --    & --    & --    & --    & 52.10 & 41.60 & 58.60 & 72.80 & --    & --    & --    & --    \\
  & AnRe   & 47.40 & 36.90 & 51.10 & 65.70 & 35.50 & 26.00 & 39.20 & 56.70 & 50.90 & 39.10 & 58.00 & 69.60 & 24.30 & 16.60 & 26.60 & 37.50 \\
  
  \midrule
  
  \multirow{13}{*}{$\spadesuit$} 
  & RE-NET & 36.93 & 26.83 & 39.51 & 54.78 & 29.78 & 19.73 & 32.55 & 48.46 & 43.67 & 33.55 & 48.83 & 62.72 & 19.55 & 12.38 & 20.80 & 34.00 \\
  & RE-GCN & 42.39 & 31.96 & 47.62 & 62.29 & 32.62 & 22.39 & 36.79 & 52.68 & 48.03 & 37.33 & 53.90 & 68.51 & 19.69 & 12.46 & 20.93 & 33.81 \\
  & RETIA & 42.76 & 32.28 & 47.77 & 62.75 & 32.43 & 22.23 & 36.48 & 52.94 & 47.26 & 36.64 & 52.90 & 67.76 & 20.12 & 12.76 & 21.45 & 34.49 \\
  & RPC & -- & -- & -- & -- & 34.91 & 24.34 & 38.74 & 55.89 & 51.14 & 39.47 & 57.11 & 71.75 & 22.41 & 14.42 & 24.36 & 38.33 \\
  & HTCCN    & 45.39 & 36.58 & 50.84 & --    & 35.63 & 24.90 & 39.26 & --    & 51.94 & 40.32 & 57.79 & --    & 23.46 & 15.18 & 25.21 & --    \\
  & TANGO    & 36.80 & 27.43 & 40.89 & 54.93 & 28.68 & 19.35 & 32.17 & 47.04 & 42.86 & 32.72 & 48.14 & 62.34 & 19.53 & 12.43 & 20.79 & 33.19 \\
  & HisMatch & 46.42 & 35.91 & 51.63 & 66.84 & 33.99 & 23.91 & 37.90 & 53.94 & 52.85 & 42.01 & 59.05 & 73.28 & 22.01 & 14.45 & 23.80 & 36.61 \\
  & CEN      & 42.20 & 32.08 & 47.46 & 61.31 & 31.50 & 21.70 & 35.44 & 50.59 & 46.84 & 36.38 & 52.45 & 67.01 & 20.39 & 12.96 & 21.77 & 34.97 \\
  & TiRGN    & 44.75 & 34.26 & 50.17 & 65.28 & 33.66 & 23.19 & 37.99 & 54.22 & 50.04 & 39.25 & 56.13 & 70.71 & 21.67 & 13.63 & 23.27 & 37.60 \\
  & LogCL    & 48.87 & 37.76 & 54.71 & 70.26 & 35.67 & 24.53 & 40.32 & 57.74 & 57.04 & 46.07 & 63.72 & 77.87 & 23.75 & 14.64 & 25.60 & 42.33 \\
  & HisRES
& \underline{50.48} & \underline{39.57} & \underline{56.65} & \underline{71.09}
& \underline{37.69} & \underline{26.46} & \underline{42.75} & \underline{59.70}
& \underline{59.07} & \underline{48.62} & \underline{65.66} & \underline{78.48}
& \underline{26.58} & \underline{16.90} & \underline{29.07} & \underline{46.31} \\

  \cmidrule(l){2-18}
  & CID-TKG  
& \textbf{51.93} & \textbf{41.49} & \textbf{57.38} & \textbf{72.32}
& \textbf{38.88} & \textbf{27.66} & \textbf{43.78} & \textbf{61.16}
& \textbf{60.44} & \textbf{50.13} & \textbf{66.93} & \textbf{79.48}
& \textbf{27.41} & \textbf{17.76} & \textbf{29.92} & \textbf{47.03} \\

  \bottomrule
  \end{tabular}
  }
  \label{tab:main_result}
\end{table*}

\subsection{Prediction and Optimization}
For link prediction, CID-TKG adopts ConvTransE following \cite{li2021temporal,chen2024local,zhang2025historically} as the scoring function to compute compatibility score between query $(e_s,r,?,t)$ and candidate entity $e_o$.
Given representations learned from either graph view $v\in\{D,I\}$, the score is defined as:
\begin{equation}
f^{v}(e_o \mid e_s, r, t)
=
\operatorname{ConvTransE}\!\left(\mathbf{e}^{v}_{s,t}, \mathbf{r}^{v}_t\right)
\cdot \mathbf{e}^{v}_{o,t},
\end{equation}
where $\mathbf{e}^{v}_{s,t}$, $\mathbf{r}^{v}_t$, and $\mathbf{e}^{v}_{o,t}$ denote the subject entity, relation, and candidate object representations, respectively.
The final prediction score is obtained by summing scores from both views:
\begin{equation}
f(e_o \mid e_s, r, t)
=
f^{D}(e_o \mid e_s, r, t)
+
f^{I}(e_o \mid e_s, r, t).
\end{equation}
A probability distribution over entities is obtained via softmax.

To enhance relation representations, we further introduce relation prediction as an auxiliary task. 
Let $\mathcal{Q}^{e}_{t}$ and $\mathcal{Q}^{r}_{t}$ denote the sets of entity and relation prediction queries at timestamp $t$, respectively. 
The training objective is defined as:
\begin{equation}
\begin{aligned}
\mathcal{L}_{\mathrm{TKG}}
=
&\;\alpha
\sum_{(e_s,r,t)\in \mathcal{Q}_t^{e}}
\mathcal{L}_{\mathrm{CE}}\!\left(e_o \mid e_s, r, t\right) \\
&\;+\;
(1-\alpha)
\sum_{(e_s,e_o,t)\in \mathcal{Q}_t^{r}}
\mathcal{L}_{\mathrm{CE}}\!\left(r \mid e_s, e_o, t\right),
\end{aligned}
\end{equation}
where $\mathcal{L}_{\mathrm{CE}}(\cdot)$ denotes the cross-entropy loss and $\alpha\in[0,1]$ balances the two tasks. 
Finally, the overall training objective combines the task loss with the contrastive alignment loss:
\begin{equation}
\mathcal{L}
=
\mathcal{L}_{\mathrm{TKG}}
+
\mu\,\mathcal{L}_{\mathrm{CoA}},
\end{equation}
where $\mu\ge0$ controls the contribution of contrastive alignment.

\subsection{Complexity Analysis}
We analyze the computational complexity of CID-TKG. 
For the spatio-temporal initialization module, the cost is dominated by the number of historical snapshots $L$, with time complexity $O\big(L (|\mathcal{E}| + |\mathcal{R}| + C_gc)\big)$, where $C_gc$ denotes the cost of graph convolution within each snapshot.
In the collaborative learning stage, the HIGE and EDGE operate on query-specific subgraphs, yielding a combined complexity of $O\big(|\mathcal{E}| + |\mathcal{F}_1| + |\mathcal{F}_2|\big)$, where $|\mathcal{F}_1|$ and $|\mathcal{F}_2|$ denote the numbers of facts in the invariance and dynamics graphs, respectively.
Overall, the computational complexity of CID-TKG is
$
O\Big(
L (|\mathcal{E}| + |\mathcal{R}| + C_gc)
+
|\mathcal{E}| + |\mathcal{F}_1| + |\mathcal{F}_2|
\Big).
$

\section{EXPERIMENTS}
\subsection{Experiment Settings}
\subsubsection{\bf Datasets.}
We evaluate our method on four widely used benchmark datasets for TKG extrapolation: ICEWS14s, ICEWS18, ICEWS05-15, and GDELT.
The first three datasets are derived from the Integrated Crisis Early Warning System (ICEWS) \cite{Boschee2015ICEWS}, while GDELT is collected from the Global Database of Events, Language, and Tone \cite{leetaru2013gdelt}.
Following prior work, all datasets are split chronologically into training,
validation, and test sets with proportions of 80\%, 10\%, and 10\%, respectively.
Detailed statistics of the datasets are reported in Table~\ref{tab:dataset_statistics_horizontal}.
Following recent extrapolation-based TKG studies \cite{zhang2025historically,chen2024local,zhang2023learning}, we do not report results on datasets with coarse temporal granularity (e.g., WIKI and YAGO), which are less suitable for modeling fine-grained temporal dynamics.

\subsubsection{\bf Evaluation Protocols.}
Following prior work \cite{li2021temporal,liang2023learn,chen2024local,zhang2025historically}, we evaluate the entity extrapolation task using
time-aware filtered metrics, including Mean Reciprocal Rank (MRR) and
Hits@$\{1,3,10\}$, and report all results in percentage form.

\subsubsection{\bf Compared Baselines.}
We select a diverse set of TKGR methods as our baselines, covering rule-based, LLM-based, and embedding-based approaches.
Specifically, the rule-based baselines include TLogic \cite{liu2022tlogic}, TR-Rules \cite{li2023tr}, and INFER \cite{li2025infer}. The LLM-based baselines consist of GenTKG \cite{liao2024gentkg}, LLM-DA \cite{wang2024large}, and AnRe \cite{tang2025anre}. The embedding-based baselines include RE-NET \cite{jin2020recurrent}, RE-GCN \cite{li2021temporal}, RETIA \cite{liu2023retia}, RPC \cite{liang2023learn}, HTCCN \cite{chen2024htccn}, TANGO \cite{han2021learning}, HisMatch \cite{li2022hismatch}, CEN \cite{li2022complex}, TiRGN \cite{li2022tirgn}, LogCL \cite{chen2024local}, and HisRES \cite{zhang2025historically}. 
Results for HGLS and L$^2$TKG are omitted due to differences in experimental settings.
All baseline results are taken from the previous works.

\subsubsection{\bf Implementation Details.}
All experiments are conducted on an NVIDIA RTX P40 GPU.
The embedding dimension is set to 200, with a dropout rate of 0.2 for all layers.
Models are trained for up to 50 epochs using early stopping based on validation MRR.
We use the Adam optimizer with a learning rate of 0.001 and a weight decay of $1\times10^{-5}$.
The historical length in the spatio-temporal initialization module is fixed to 3.
The GNN encoder consists of two layers.
Hyperparameters, including the contrastive alignment weight $\mu$, temperature $\gamma$, number of GAT layers, and rule graph length, are tuned via grid search on the validation set.
For ICEWS14s, ICEWS18, ICEWS05-15, and GDELT, $\mu$ is set to 0.2, 0.01, 0.15, and 0.3, $\gamma$ to 0.3, 0.03, 0.2, and 0.25, the number of GAT layers to 3, 2, 2, and 2, and the rule graph length to 10, 8, 10, and 8, respectively.

\subsection{Main Results}

Table~\ref{tab:main_result} reports the time-aware filtered evaluation results for the entity extrapolation task on four benchmark datasets.
Overall, CID-TKG consistently outperforms most strong baselines on all benchmarks.
Specifically, CID-TKG achieves clear advantages over rule-based and LLM-based approaches, suggesting improved robustness to variations in data distributions.
Compared with embedding-based methods, CID-TKG also delivers superior performance on all datasets.
In particular, relative to the second-best results, CID-TKG obtains substantial MRR improvements of 2.9\%, 3.2\%, 2.3\%, and 3.1\% on ICEWS14s, ICEWS18, ICEWS05-15, and GDELT, respectively. 
These consistent gains indicate that CID-TKG learns more effective representations by jointly modeling evolutionary dynamics and historical structure, thereby improving prediction accuracy and generalization ability.
Overall, the results verify the effectiveness of our CID-TKG for TKG extrapolation.

\begin{table}[h]
  \centering
  \footnotesize
  \setlength{\tabcolsep}{4pt}
  \caption{Ablation results on ICEWS14s and ICEWS18.}
  \resizebox{0.47\textwidth}{!}{
  \begin{tabular}{lcccccc}
  \toprule
  \multirow{2}{*}{Variant} 
  & \multicolumn{3}{c}{ICEWS14s} 
  & \multicolumn{3}{c}{ICEWS18} \\
  \cmidrule(lr){2-4}\cmidrule(lr){5-7}
  & MRR & Hits@1 & Hits@10 
  & MRR & Hits@1 & Hits@10 \\
  \midrule

  CID-TKG
  & \textbf{51.93} & \textbf{41.49} & \textbf{72.32}
  & \textbf{38.88} & \textbf{27.66} & \textbf{61.16} \\
  \midrule
  
  -w/o-$\mathcal{G}^{D}$
  & 49.54 & 38.82 & 70.00
  & 36.88 & 25.53 & 59.40 \\
  -w/o-$\mathcal{G}^{I}$
  & 50.29 & 39.75 & 70.55
  & 36.98 & 25.81 & 58.92 \\
  -w/o-(CoA \& ReD)
  & 50.96 & 40.44 & 71.10
  & 38.42 & 27.15 & 60.80 \\
  -w/o-CoA
  & 51.37 & 41.04 & 71.15
  & 38.62 & 27.46 & 60.57 \\
  -w/o-ReD
  & 51.48 & 40.92 & 71.52
  & 38.47 & 27.23 & 60.62 \\
  \midrule
  
  -w/o-TE
  & 51.35 & 40.77 & 71.68
  & 38.55 & 27.26 & 61.09 \\
  -w/cos
  & 51.38 & 41.01 & 71.51
  & 38.60 & 27.32 & 60.94 \\
  \midrule
  -w/EnD
  & 51.76 & 41.24 & 71.62
  & 38.39 & 27.14 & 60.78 \\
  -w/Simp
  & 44.91 & 34.17 & 66.32
  & 33.82 & 23.05 & 55.10 \\

  \bottomrule
  \end{tabular}
  }
  \label{tab:ablation}
\end{table}
\subsection{Ablation Study}

We conduct ablation studies on ICEWS14s and ICEWS18 to assess the contributions of components in CID-TKG by MRR and Hits@1,10. 
The results are summarized in Table~\ref{tab:ablation}, where nine variants are evaluated: 
(1) the full CID-TKG;
(2) without the evolutionary dynamics graph (``-w/o-$\mathcal{G}^{D}$'');
(3) without the historical invariance graph (``-w/o-$\mathcal{G}^{I}$'');
(4) without both relation decomposition and contrastive alignment (``-w/o-(CoA \& ReD)'');
(5) without contrastive alignment (``-w/o-CoA'');
(6) without relation decomposition (``-w/o-ReD'');
(7) without the time encoder in the evolutionary dynamics encoder (``-w/o-TE'');
(8) replacing the proposed time encoder with cosine encoding (``-w/cos'');
(9) introducing entity decomposition analogous to relation decomposition (``-w/EnD'') and (10) replacing the rule--based retrieval strategy with a simple heuristic that constructs $\mathcal{G}^{D}$ (``-w/Simp'').

Overall, removing either the historical invariance graph or the evolutionary dynamics graph results in substantial performance degradation, whereas jointly modeling both consistently yields the best results, validating the effectiveness of the proposed collaborative learning framework. 
Excluding relation decomposition or contrastive alignment also results in clear performance drops, highlighting the importance of view-specific relation representations and cross-view alignment for suppressing view-specific noise.
We further analyze temporal encoding strategies in the evolutionary dynamics encoder. 
Removing time encoder significantly degrades performance, and replacing the proposed encoder with cosine encoding yields inferior results, demonstrating the importance of explicitly modeling temporal relevance for extrapolation.
Applying a decomposition mechanism to entity representations also leads to slight performance degradation, suggesting that relation semantics benefit more from explicit decomposition, whereas entity decomposition may disrupt shared contextual information.
In addition, we evaluate a rule-free heuristic for constructing $\mathcal{G}^{D}$ (``-w/Simp'') by selecting the $N$ most recent facts involving the query subject. 
Compared with ``-w/o-$\mathcal{G}^{I}$'', this naive retrieval strategy incurs a substantial performance drop, indicating that simple temporal retrieval is insufficient for capturing informative evolutionary dynamics.

\begin{figure}[t]
  \centering
  \subcaptionbox{ICEWS14s\label{fig:left_length}}[0.48\linewidth]{%
    \includegraphics[width=\linewidth]{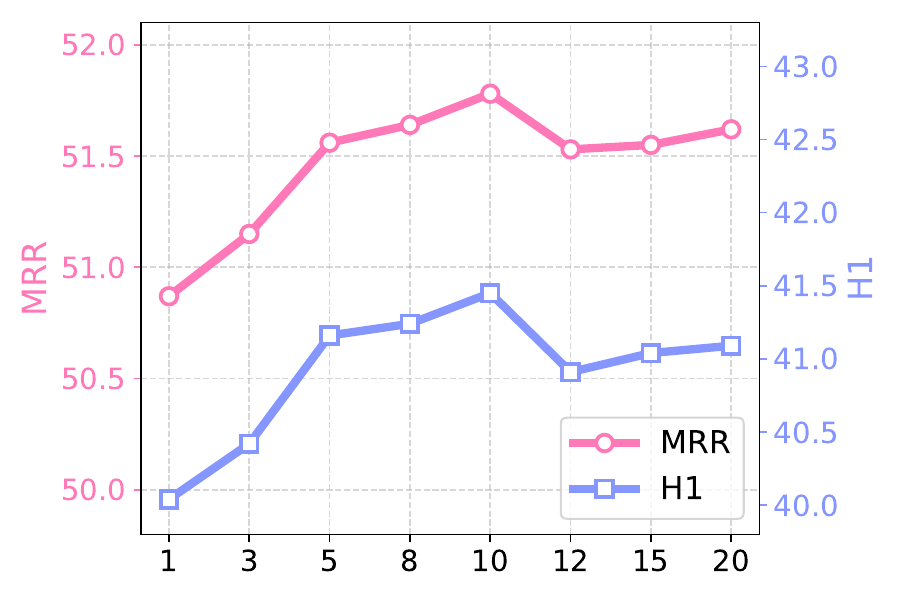}
  }\hspace{0.0\linewidth}
  \subcaptionbox{ICEWS18\label{fig:right_length}}[0.48\linewidth]{%
    \includegraphics[width=\linewidth]{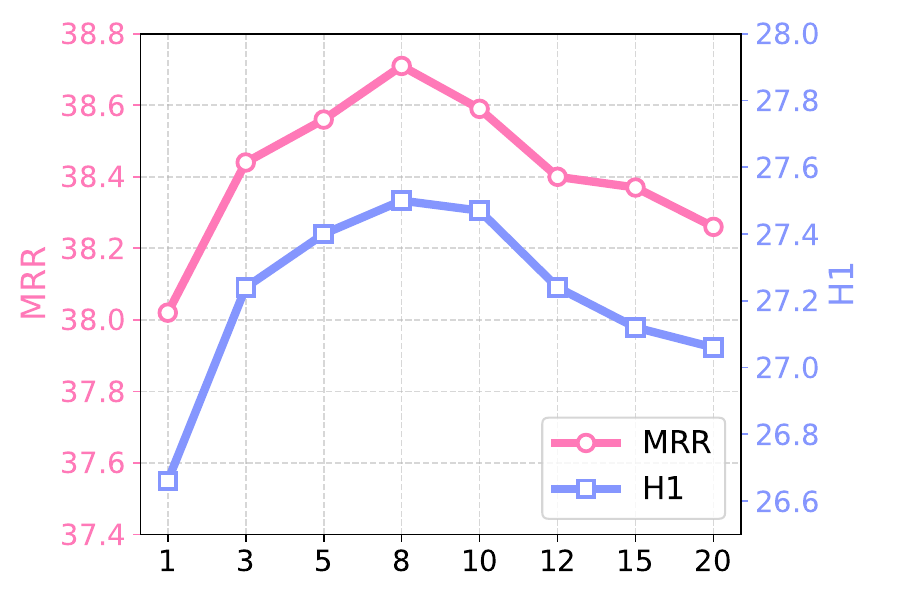}
  }
  \vspace{-5pt}
  \caption{Effect of the upper bound $N$ on the evolutionary dynamics graph.}
  \Description{Effect of the upper bound $N$ on the evolutionary dynamics graph.}
  \label{fig:overall_length}
\end{figure}

\begin{figure}[t]
  \centering
  \subcaptionbox{ICEWS14s\label{fig:left_mu}}[0.48\linewidth]{%
    \includegraphics[width=\linewidth]{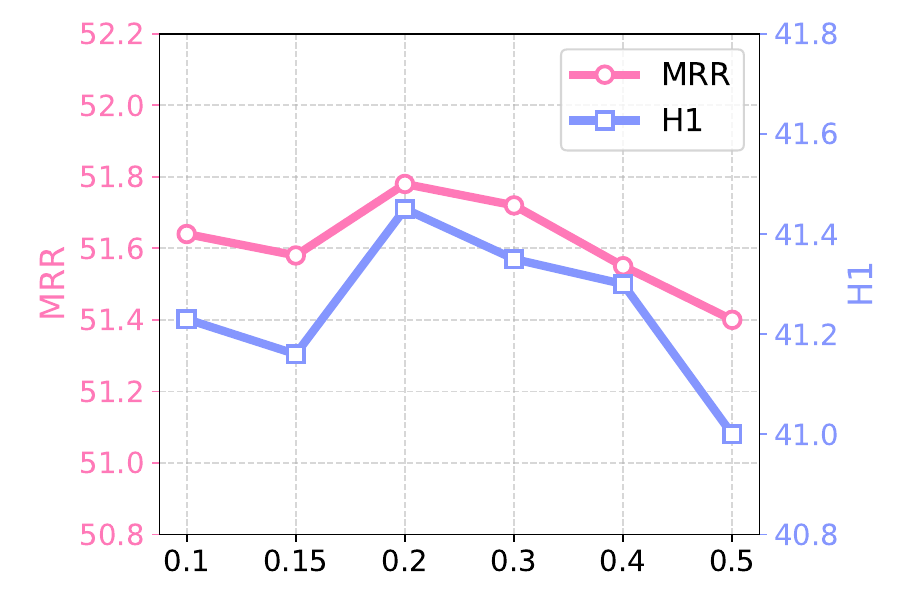}
  }\hspace{0.0\linewidth}
  \subcaptionbox{ICEWS18\label{fig:right_mu}}[0.48\linewidth]{%
    \includegraphics[width=\linewidth]{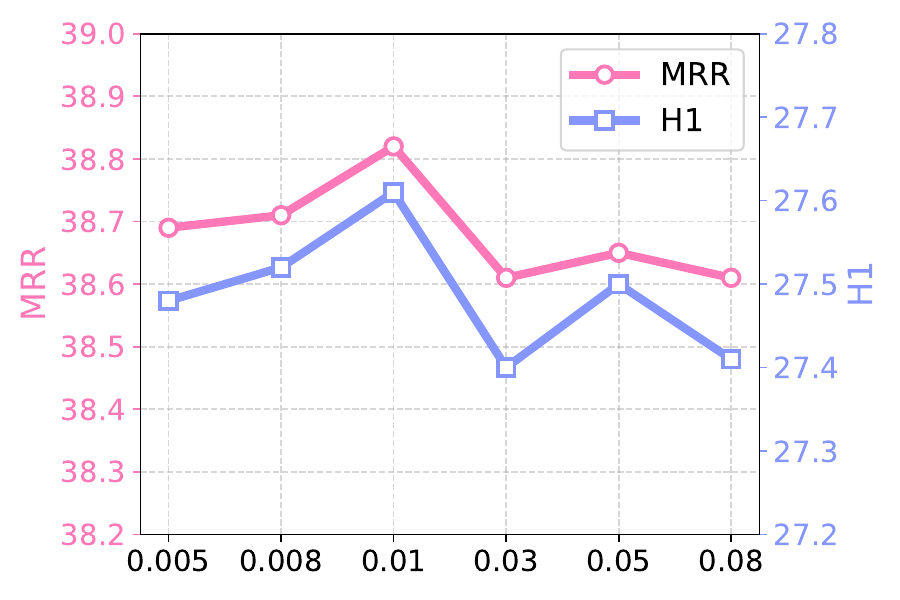}
  }
  \vspace{-5pt}
  \caption{Effect of the contrastive alignment weight $\mu$.}
  \Description{Effect of the contrastive alignment weight $\mu$.}
  \label{fig:overall_mu}
\end{figure}

\begin{table*}[ht!]
  \centering
  \caption{Case study of CID-TKG's entity extrapolation performance on the ICEWS14s dataset. The ground-truth entity is highlighted in bold, and the corresponding prediction scores are provided.}
  \footnotesize
  \resizebox{0.98\textwidth}{!}{
  \begin{tabular}{l cc cc cc}
  \toprule
  \multirow{2}{*}{Query} 
    & \multicolumn{2}{c}{CID-TKG-w/$\mathcal{G}^D$} 
    & \multicolumn{2}{c}{CID-TKG-w/$\mathcal{G}^I$}
    & \multicolumn{2}{c}{CID-TKG} \\
  \cmidrule(lr){2-3}\cmidrule(lr){4-5}\cmidrule(lr){6-7}
    & Hits@5 Entity & Score & Hits@5 Entity & Score & Hits@5 Entity & Score \\
  \midrule
  
  \multirow{5}{*}{\makecell[l]{
    \makecell[l]{$e_s:Head\_of$ \\ $\_Government\_(India)$} \\
    $r:Praise\_or\_endorse$ \\
    $t:2014-12-01$ }}
    & $Member\_of\_Parliament\_(India)$ & 5.1521 & $Citizen\_(India)$ & 4.6465 & $\bm{Employee\_(India)}$ & \textbf{9.5145} \\
    & $Governor\_(India)$               & 5.1514 & \makecell[c]{$Company\_-\_Owner$ \\ $\_or\_Operator\_(India)$} & 4.4339 & \makecell[c]{$Member \_of$ \\ $\_Parliament\_(India)$} & 9.1098 \\
    & $\bm{Employee\_(India)}$          & \textbf{5.1435} & $Indigenous\_People\_(India)$ & 4.4003 & $Governor\_(India)$ & 8.9817 \\
    & $Labor\_Union\_(India)$           & 4.4350 & $\bm{Employee\_(India)}$ & \textbf{4.3710} & $Citizen\_(India)$ & 8.7354 \\
    & $Administrative\_Body\_(India)$   & 4.3371 & $Villager\_(India)$ & 4.2844 & $Labor\_Union\_(India)$ & 7.9872 \\
  \midrule
  
  \multirow{5}{*}{\makecell[l]{
    $e_s:Angola$ \\
    $r:Host\_a\_visit$ \\
    $t:2014-12-02$ }}
    & $Cape\_Verde$                   & 9.3489 & $William\_Ruto$ & 6.9054 & $\bm{Foreign\_Affairs\_(Portugal)}$ & \textbf{14.5256} \\
    & $\bm{Foreign\_Affairs\_(Portugal)}$  & \textbf{8.5451} & \makecell[c]{$Government\_Official$ \\ $\_(United\_States)$} & 6.5515 & $William\_Ruto$ & 13.8299 \\
    & $Nonofo\_Molefhi$               & 7.8376 & $Barack\_Obama$ & 6.3843 & \makecell[c]{$Government\_Official$ \\ $\_(United\_States)$} & 13.7910 \\
    & $Louise\_Mushikiwabo$           & 7.7446 & $Sergey\_Viktorovich\_Lavrov$ & 6.0563 & $Gen\_Prayut\_Chan-o-cha$ & 12.6733 \\
    & \makecell[c]{$National\_Security$\\ $\_Advisor\_(Thailand)$} & 7.7324 & $\bm{Foreign\_Affairs\_(Portugal)}$ & \textbf{5.9805} & $Sergey\_Viktorovich\_Lavrov$ & 12.2393 \\
  \midrule
  
  \multirow{5}{*}{\makecell[l]{
    $e_s:Afghanistan$ \\
    \makecell[l]{$r:Engage\_in$ \\ $\_negotiation$} \\
    $t:2014-12-25$ }}
    & $Abdullah\_Abdullah$       & 8.0501 & $Iraq$ & 10.2351 & $\bm{Tajikistan}$ & \textbf{16.7835} \\
    & $Raheel\_Sharif$           & 8.0464 & $\bm{Tajikistan}$ & \textbf{9.9402} & $Iraq$ & 15.1064 \\
    & $China$                    & 7.3678 & $Afghanistan$ & 9.1128 & $China$ & 14.8508 \\
    & $Refugee\_(Afghanistan)$   & 7.0583 & $Iran$ & 7.9106 & $Iran$ & 14.1855 \\
    & $\bm{Tajikistan}$          & \textbf{6.8432} & $China$ & 7.4830 & $Afghanistan$ & 12.9165 \\
  
  \bottomrule
  \end{tabular}}
  \label{tab:dual_hits5}
\end{table*}

\subsection{Sensitivity Analysis}\label{sec:sensitive}

We experiment on ICEWS14s and ICEWS18 to analyze the effects of key hyperparameters in CID-TKG, including the upper bound $N$ on the number of historical facts for constructing the evolutionary dynamics graph and the contrastive alignment weight $\mu$. Fig.~\ref{fig:overall_length} shows the impact of $N$ on performance. The best results are achieved at $N=10$ on ICEWS14s and $N=8$ on ICEWS18. 
When $N$ is too small, limited historical information hinders accurate prediction, while overly large $N$ introduces noisy histories and increases computational cost. The smaller optimal $N$ on ICEWS18 can be attributed to its denser event distribution and more stable evolutionary patterns.
Fig.~\ref{fig:overall_mu} illustrates the effect of $\mu$, which controls contrastive alignment strength. Optimal performance is obtained at $\mu=0.2$ on ICEWS14s and $\mu=0.01$ on ICEWS18, indicating that the optimal alignment strength is dataset-dependent. 
In particular, ICEWS14s benefits from stronger alignment due to sparser events and less stable dynamics, whereas ICEWS18 requires weaker alignment given its denser and more stable temporal structure.

\begin{figure}[h]
  \centering
  \subcaptionbox{ICEWS14s\label{fig:left_robust}}[0.48\linewidth]{%
    \includegraphics[width=\linewidth]{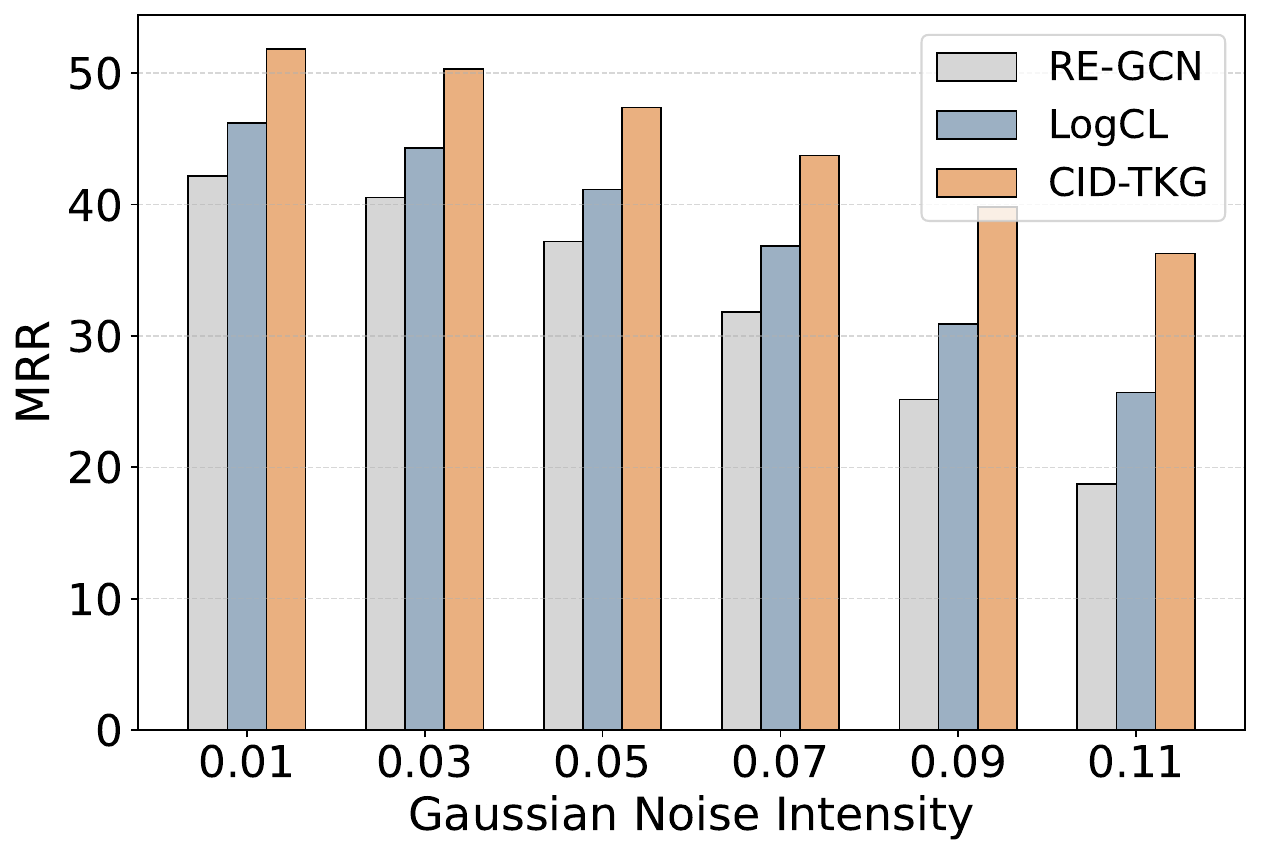}
  }\hspace{0.0\linewidth}
  \subcaptionbox{ICEWS18\label{fig:right_robust}}[0.48\linewidth]{%
    \includegraphics[width=\linewidth]{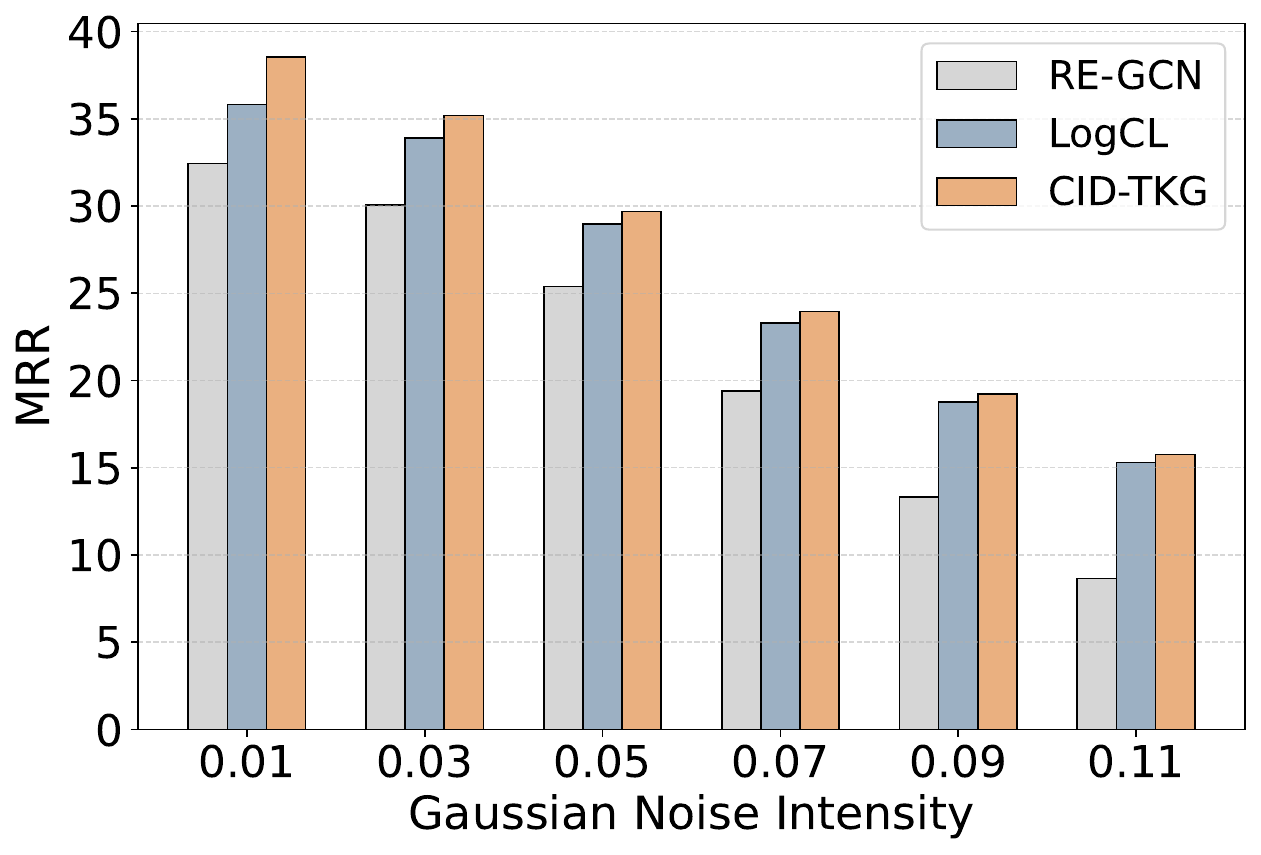}
  }
  \vspace{-5pt}
  \caption{Robustness analysis under Gaussian noise perturbations.}
  \Description{Robustness analysis under Gaussian noise perturbations.}
  \label{fig:overall_robust}
\end{figure}

\subsection{Robustness Analysis}

We evaluate robustness by injecting Gaussian noise with varying intensities into the initial entity embeddings at inference time. 
We compare CID-TKG with RE-GCN \cite{li2021temporal} and LogCL \cite{chen2024local}. The results are shown in Fig.~\ref{fig:overall_robust}.
On ICEWS14s, CID-TKG demonstrates the strongest robustness, exhibiting substantially smaller performance degradation as noise intensity increases. 
When the noise level reaches 0.11, CID-TKG experiences an MRR drop of 30.04\%, compared to 55.54\% for RE-GCN and 44.37\% for LogCL, indicating greater resilience to perturbations in entity representations.
As analyzed in Section~\ref{sec:sensitive}, the evolutionary dynamics in ICEWS18 are relatively more stable, leading to smaller marginal robustness gains of CID-TKG than those observed on ICEWS14s.
However, CID-TKG still has stronger robustness than other baselines on ICEWS18.

\begin{figure}[t]
  \centering
  \subcaptionbox{ICEWS14s\label{fig:left_time}}[0.48\linewidth]{%
    \includegraphics[width=\linewidth]{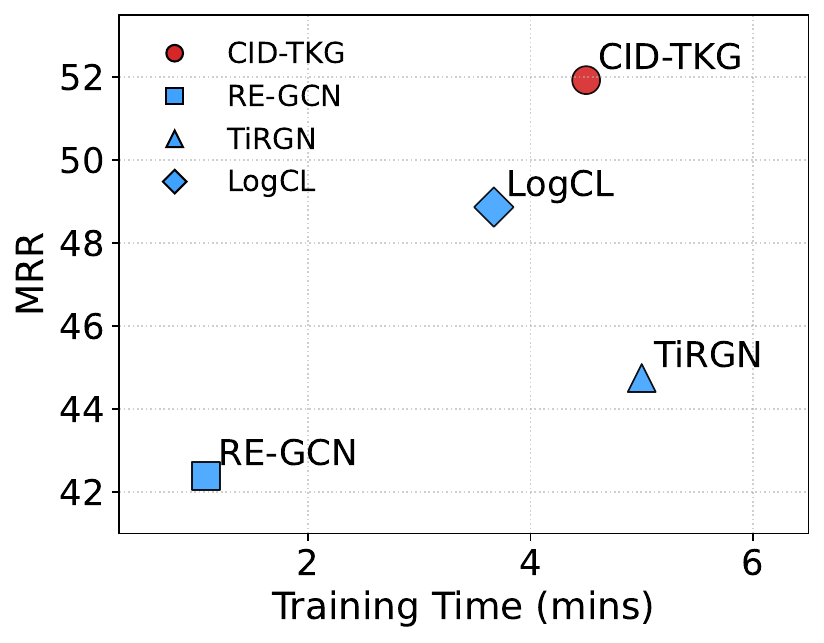}
  }\hspace{0.02\linewidth}
  \subcaptionbox{ICEWS18\label{fig:right_time}}[0.48\linewidth]{%
    \includegraphics[width=\linewidth]{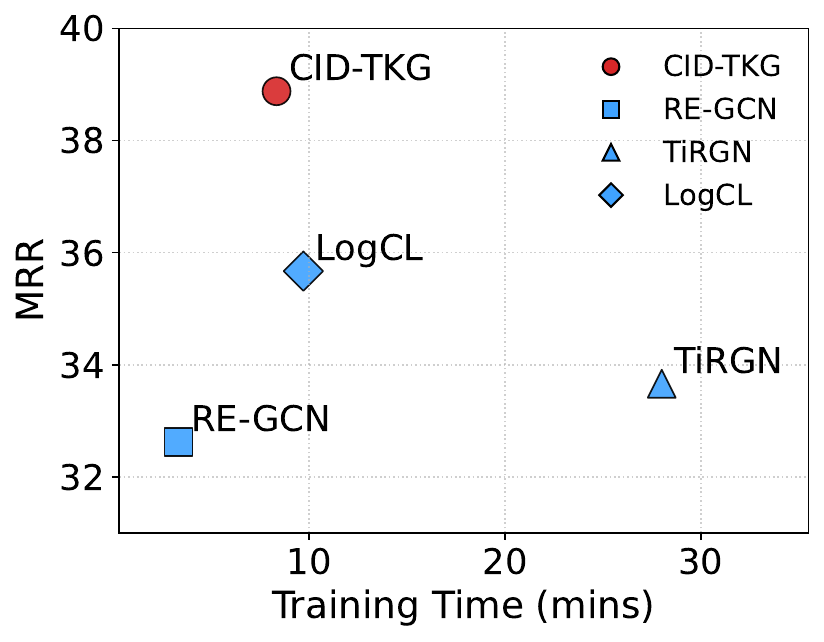}
  }
  \vspace{-5pt}
  \caption{Efficiency–performance trade-off comparisons.}
  \Description{Efficiency–performance trade-off of different methods.
  }
  \label{fig:overall_time}
\end{figure}

\subsection{Execution Time Analysis}

We analyze the efficiency and predictive performance of our CID-TKG against strong baselines, including RE-GCN \cite{li2021temporal}, TiRGN \cite{li2022tirgn}, and LogCL \cite{chen2024local}, as shown in 
Fig.~\ref{fig:overall_time}.
On both datasets, CID-TKG achieves a favorable balance between efficiency and performance. 
Compared with RE-GCN, CID-TKG incurs moderately higher training time but consistently yields substantially better MRR. 
In contrast, more complex methods such as TiRGN and LogCL exhibit significantly higher training costs but underperform CID-TKG.
These results demonstrate that CID-TKG provides an effective trade-off between computational efficiency and extrapolation accuracy. 
Despite encoding both historical invariance and evolutionary dynamics graphs, efficient spatio-temporal initialization with a limited number of historical snapshots enables CID-TKG to achieve strong performance without excessive overhead.
In addition, CID-TKG incurs only slightly higher testing time than the simple baseline RE-GCN, but more efficient than TiRGN and LogCL across ICEWS14s, ICEWS18, ICEWS05-15, and GDELT.

\subsection{Comparison with Temporal Modeling Baselines}

To provide a more comprehensive evaluation, we further compare CID-TKG with several representative time-aware interpolation-based methods, including TTransE \cite{leblay2018deriving}, TA-DistMult \cite{garcia2018learning}, DE-SimplE \cite{goel2020diachronic}, TNTComplEX \cite{lacroixtensor}, and ECEformer \cite{fang2024transformer}.
All methods are evaluated on extrapolation datasets using time-aware filtered metrics, and the results are reported in Table~\ref{tab:addition}.
Among these baselines, ECEformer achieves the strongest overall performance.
However, similar to other interpolation-based approaches, ECEformer relies on explicit timestamp representations that must be observed during training.
This design limits its applicability in extrapolation settings, where predictions are required at unseen future timestamps.
These results further demonstrate the advantage of CID-TKG in the entity extrapolation task, highlighting its ability to effectively model future temporal dynamics beyond the scope of interpolation-based methods.

\begin{table}[t]
  \centering
  \footnotesize
  \caption{Additional Comparison with Temporal Modeling Baselines on ICEWS14s and ICEWS18.}
  \setlength{\tabcolsep}{4pt}
  \resizebox{0.47\textwidth}{!}{
  \begin{tabular}{lcccccc}
  \toprule
  \multirow{2}{*}{Variant} 
  & \multicolumn{3}{c}{ICEWS14s} 
  & \multicolumn{3}{c}{ICEWS18} \\
  \cmidrule(lr){2-4}\cmidrule(lr){5-7}
  & MRR & Hits@1 & Hits@10 
  & MRR & Hits@1 & Hits@10 \\
  \midrule
  
  TTransE
  & 13.72 & 2.98 & 35.74
  & 8.31 & 1.92 & 21.89 \\
  TA-DisMult 
  & 25.80 & 16.94 & 42.99
  & 16.75 & 8.61 & 33.59 \\
  DE-simlE
  & 33.36 & 24.85 & 49.82
  & 19.30 & 11.53 & 34.80 \\
  TNTComplEX 
  & 34.05 & 24.85 & 49.82
  & 21.23 & 13.28 & 36.91 \\
  ECEformer
  & 34.84 & 25.60 & 52.67
  & 26.01 & 16.67 & 44.33 \\
  
  \midrule
  CID-TKG
  & \textbf{51.93} & \textbf{41.49} & \textbf{72.32}
  & \textbf{38.88} & \textbf{27.66} & \textbf{61.16} \\
  \bottomrule
  \end{tabular}
  }
  \label{tab:addition}
\end{table}

\subsection{Case Study}

To further demonstrate the effectiveness of CID-TKG, we conduct a qualitative analysis on several representative facts from the ICEWS14s dataset using different model variants.
The results are reported in Table~\ref{tab:dual_hits5}, where “CID-TKG-w/$\mathcal{G}^{D}$” and “CID-TKG-w/$\mathcal{G}^{I}$” denote variants that rely solely on the evolutionary dynamics graph and the historical invariance graph, respectively, and “CID-TKG” represents the full model.
The ground-truth entity for each query is highlighted in bold, and the top-ranked entities under the Hits@5 setting along with their scores are reported.
From the first example, the ground-truth entity \textit{Employee (India)} appears in the top-5 predictions when using either graph alone, but is not ranked at the top by either variant.
By combining scores from both graph views, CID-TKG assigns the highest score to the ground-truth entity, leading to a correct prediction. 
These cases demonstrate that the historical invariance graph and the evolutionary dynamics graph provide complementary evidence.
Similar observations can be made for the other examples.
Jointly modeling both views enables more accurate and reliable entity extrapolation, highlighting the effectiveness of the proposed collaborative learning framework.

\section{Conclusion}
In this work, we propose CID-TKG, a collaborative learning framework that jointly models historical invariance structures and evolutionary dynamics patterns through two complementary graph views. 
These two views are encoded by a Historical Invariance Graph Encoder and an Evolutionary Dynamics Graph Encoder, respectively,
to capture stable structural semantics and query-specific temporal dynamics.
By introducing relation decomposition and contrastive alignment, CID-TKG effectively mitigates semantic conflicts and view-specific noise in collaborative learning, leading to more robust and informative representations.
We further provided an information-theoretic analysis that offers principled insights into why collaborative learning improves extrapolation performance.
Extensive experiments on multiple benchmark datasets demonstrated that CID-TKG consistently outperforms state-of-the-art methods under extrapolation settings.
Together, these results validate the effectiveness of jointly modeling historical invariance and evolutionary dynamics structures for TKGR.

\bibliographystyle{ACM-Reference-Format}
\bibliography{sample-base}

\end{document}